\documentclass[11pt,a4paper]{article}
\usepackage[utf8]{inputenc}
\usepackage[T1]{fontenc}
\usepackage{amsmath,amssymb}
\usepackage[round,authoryear]{natbib}
\usepackage{graphicx}
\usepackage{booktabs}
\usepackage{tabularx}
\usepackage{array}
\usepackage{longtable}
\usepackage{hyperref}
\usepackage{url}
\usepackage{listings}
\usepackage{xcolor}
\usepackage{microtype}
\usepackage{parskip}
\usepackage{CJKutf8}
\newcommand{\cn}[1]{\begin{CJK*}{UTF8}{gbsn}#1\end{CJK*}}

\newcolumntype{Y}{>{\raggedright\arraybackslash}X}
\newcolumntype{L}[1]{>{\raggedright\arraybackslash}p{#1}}
\newcolumntype{C}[1]{>{\centering\arraybackslash}p{#1}}

\hypersetup{
    colorlinks=true,
    linkcolor=blue,
    citecolor=blue,
    urlcolor=blue,
    pdftitle={Traceable Scholarship: Page Anchors and Ariadne's Thread for Humanistic Inquiry in the Age of Generative AI},
    pdfauthor={Deyu Jing},
}

\title{\textbf{Traceable Scholarship: Page Anchors and Ariadne's Thread for Humanistic Inquiry in the Age of Generative AI}}
\author{Deyu Jing\\
        Fudan University, Shanghai, China\\
        \texttt{baireinhold@163.com}}
\date{September 5, 2026}

\begin{document}

\maketitle

\begin{abstract}
Generative AI lets large language models produce scholarly-looking text within seconds, yet fluency does not equal valid explanation. The deepest risk is not factual error alone but the \textit{appearance} that an explanation is already established without clear sources, page numbers, editions, or evidence. We liken the page anchor to \textbf{Ariadne's thread}: within the labyrinth of generative fluency, it is the thread that leads the scholar back to the source. This paper proposes \textbf{Traceable Scholarship} as the minimum normative condition for AI-assisted humanistic research, situating it across the three revolutions of knowledge infrastructure---print, digital, and generative AI. We introduce page anchors, dual page numbers, citation-first generation, NO\_EVIDENCE, and human verification, four-level compliance, and Scope Contract, and present \textbf{AIH-Infra} as a three-layer reference implementation: Contexture (document structuring), Open WebUI AIH-Infra (traceable knowledge base), and AIH-Infra MCP Server (agent gateway). A case study on a 29-volume Kant \textit{Akademie-Ausgabe} knowledge base illustrates how traceability supports retrieval correction, evidence grading, and judgment downgrading. Traceability is not a software feature; it is the condition under which humanistic research can remain public and refutable in the age of generative AI.
\end{abstract}


\section{Introduction: Three Revolutions of Knowledge Infrastructure and the Crisis of Traceability}\label{sec:intro}

The third revolution of knowledge infrastructure is unfolding faster than our ability to comprehend it.

The first revolution was \textbf{print}, which gave texts \textbf{reproducibility}. This revolution unfolded almost simultaneously across Eurasia: in Europe, Gutenberg's movable type gave rise to the footnote system---a social contract anchoring scholarly claims to verifiable sources \citep{grafton1997footnote}; in China, the woodblock and movable-type traditions that developed from Bi Sheng onward were deeply embedded in the systematic compilation of \textit{leishu} encyclopedias and \textit{congshu} collectanea, as well as in the four-part bibliographic system, cultivating a highly developed textual consciousness of collation and edition sensitivity. The two civilizational paths differ, yet they share one core fact: print changed not the speed of writing but the \textbf{verifiability and transmissibility of knowledge} \citep{eisenstein1979printing}.

The second revolution was \textbf{digitization}, which gave texts \textbf{searchability}. From the 1990s onward, full-text search made vast corpora navigable; JSTOR, Google Scholar, and various digital archives greatly expanded scholarly accessibility. Yet searchability did not abolish the requirement of traceability; it merely shifted the medium of tracing from paper pages to digital locators.

The third revolution is \textbf{generative AI}, which gives texts \textbf{generability}. Large language models allow anyone to obtain fluent, seemingly authoritative scholarly expositions within seconds. But this revolution introduces an unprecedented rupture: \textbf{generated text can exist without relying on any source at all}. In the first two revolutions, traceability remained an invariant underlying principle---printed books had page numbers, digital documents had URLs and DOIs---but generative AI outputs naturally lack source anchors unless we actively embed them.

This historical pattern is not our invention. As early as 1792, Herder recognized the logic of intergenerational leaps in cognitive infrastructure: later generations understand the classics better than their predecessors not because individuals are smarter, but because ``we stand on their shoulders\ldots\ gaining new perspectives they could not have and discovering new sources they could not know'' \citep{herder1792briefe}. His core insight---that the progress of understanding depends on the progress of infrastructure rather than on individual intelligence---has been confirmed again and again across the three revolutions. The invariant across all three is traceability. This paper argues that in the third revolution, traceability has not become obsolete; it is more urgent than ever.

The most common concern in discussions of generative AI and academic research is \textbf{hallucination}: models may fabricate facts, forge references, misread materials, or cite non-existent sources \citep{bender2021dangers,ji2023survey}. This risk is real, but framing the problem merely as ``models make mistakes'' underestimates the structural impact of generative AI on humanistic research.

In the humanistic context, the deeper risk is that even when a model does not obviously fabricate, it can organize locally relevant, locally valid, and locally questionable materials into a seemingly complete explanation. It can generate coherent paragraphs, appropriate terminology, measured tone, and apparently mature argumentative structure, giving the researcher the impression that the explanation has already been formed and the judgment is sufficient, with only polishing and footnote-filling remaining. In other words, generative AI does not only produce errors; it produces the \textbf{appearance that an explanation has already been established} (\cn{解释外观}).

This appearance is especially dangerous for the humanities. Humanistic research is not merely the organization of information into fluent text. Judgments in history, philosophy, literature, intellectual history, classics, and philology typically depend on edition, page, paragraph, footnote, marginal note, context, counter-example, and historiographical location. Whether a judgment holds cannot be decided by how smoothly it is stated; it must be traceable, questionable, reviewable, and refutable.

The footnote, edition statement, collation note, archive number, and page citation in traditional scholarly institutions exist precisely so that judgments do not remain within the author's linguistic authority but can be returned to the materials themselves \citep{grafton1997footnote}. The problem with generative AI is that it can bypass these institutional mediators and directly output a text with scholarly appearance. If researchers merely append sources after the generation without checking the real relationship between claim and source, the phenomenon we call \textbf{hallucination laundering} (\cn{幻觉洗白}) occurs: sourceless or weakly sourced AI-generated judgments, packaged through citation format, acquire the appearance of scholarly authority.

Therefore, the fundamental question of AI-assisted humanistic research should move from ``how to reduce hallucination'' to ``how to prevent the appearance of explanation from being established before the evidential conditions are met.'' Our answer is: we must establish a minimum norm for Traceable Scholarship (\cn{可溯源学术研究}), so that any AI-assisted judgment entering scholarly use can be traced back to specific materials, specific pages, specific editions, and reviewable research processes.

The urgency of this problem has risen rapidly in recent years. On the one hand, RAG research has broadly addressed external knowledge retrieval, answer faithfulness, and citation generation \citep{lewis2020retrieval}, and research on AI hallucinations has systematically mapped the mechanisms by which language models generate unreliable content \citep{ji2023survey}. On the other hand, digital humanities has long dealt with issues of edition, provenance, structure, and infrastructural power \citep{tei2023p5,nockels2022transkribus,zaagsma2023digital}. Yet existing work often disperses ``source,'' ``page,'' and ``evidence chain'' (\cn{证据链}/\cn{证据回链}) across different technical stages or disciplinary traditions, without forming a unified normative-infrastructural framework for AI-assisted humanistic research. The author publicly released an early working draft on Zenodo in February 2026, first proposing Traceable Scholarship, page anchors (\cn{页面锚点}/\cn{页码锚点}), NO\_EVIDENCE (\cn{当前作用域内无足够证据}), and the four-level compliance framework \citep{jing2026zenodo}. Building on that conceptual framework, this paper integrates the April 2026 Kant case and the June--July 2026 developments in the three-layer AIH-Infra architecture and MCP integration, advancing the abstract norm toward a runnable infrastructure account.

This paper makes three main contributions. First, at the normative level, it proposes an operational definition of Traceable Scholarship and distinguishes the explanatory responsibilities of source-dependent, synthetic-interpretive, and normative-theoretical claims. Second, at the infrastructure level, it proposes a capability framework composed of page anchors, dual page numbers (\cn{双重页码}), citation-first generation (\cn{引文优先生成}; \cn{证据先行}), NO\_EVIDENCE, four-level compliance, and Scope Contract (\cn{作用域契约}; \cn{研究范围契约}), and presents AIH-Infra's three-layer architecture as a reference implementation. Third, at the research-practice level, through the case of a 29-volume Kant \textit{Akademie-Ausgabe} knowledge base, it shows how traceable infrastructure supports retrieval-strategy correction, material-layer distinction, evidence grading, and judgment downgrading (\cn{判断降调}), while noting that this demonstrative case cannot substitute for cross-case or third-party evaluation.

\section{Defining Traceable Scholarship}\label{sec:defining}

This paper defines \textbf{Traceable Scholarship} as follows:

\begin{quote}
In AI-assisted humanistic research based on documents, archives, or other locatable materials, source-dependent claims must be traceable to clear material boundaries, editions, locators, and evidence fragments; synthetic interpretive claims must expose their evidence sets, inferential layers, and necessary processing steps; when the system cannot find sufficient evidence within the current scope, it must explicitly acknowledge evidential insufficiency rather than disguise model background knowledge as material conclusion.
\end{quote}

This definition contains five layers.

First, \textbf{source boundaries}. The system must indicate which knowledge bases, files, editions, and retrieval ranges the current judgment is based on, rather than conflating model world knowledge, public web materials, and local document collections.

Second, \textbf{stable locators}. The system should connect judgments to page numbers, volume numbers, folio numbers, Stephanus codes, Bekker codes, archive numbers, image coordinates, time codes, and other stable locators whenever possible. For modern books and PDF documents, the page anchor is the most basic and commonly used locating mechanism.

Third, \textbf{evidence fragments}. A citation must not stop at ``from a certain file'' or ``according to a certain book.'' The system should expose the specific textual fragments, context windows, chapter paths, and necessary page information that support the judgment.

Fourth, \textbf{process records}. Retrieval parameters, query expressions, recalled results, reranking strategies, reading windows, and excluded materials may all affect the final judgment. Traceable research does not require every step to be infinitely expanded, but it should at least preserve sufficient Evidence/Trace to reconstruct key judgments.

Fifth, \textbf{no-evidence discipline}. When the system finds no evidence matching the question, it should output NO\_EVIDENCE or an equivalent marker rather than using the model's own knowledge to fill in a fluent answer. Silence, downgrade, and pause are part of scholarly honesty.

Traceable Scholarship does not require AI systems to complete scholarly judgment automatically. On the contrary, it recognizes that AI can only provide candidate evidence, candidate interpretations, candidate connections, and candidate questions. What it truly requires is that these candidates be re-checkable by researchers, reviewers, readers, and the scholarly community. Judgments that cannot be returned to source locations are, at most, research clues, not scholarly judgments that can directly enter a paper.

\subsection{Three Types of Claims and Their Distinct Traceability Requirements}\label{sec:claims}

To avoid reducing humanistic interpretation to mechanical citation, this paper distinguishes three types of claims (Table~\ref{tab:claims}).

\begin{table}[htbp]
\centering
\caption{Three Types of Claims and Their Traceability Requirements}
\label{tab:claims}
\small
\begin{tabularx}{\textwidth}{@{}L{.20\textwidth}YY@{}}
\toprule
\textbf{Claim Type} & \textbf{Typical Form} & \textbf{Traceability Requirement} \\
\midrule
Source-dependent claim & ``Kant divides humankind into several races in a given text.'' & Must return to the original text, edition, and stable locator; distinguish direct quotation from paraphrase. \\
\addlinespace
Synthetic interpretive claim & ``There is a structural tension between Kant's universal history and his racial classification.'' & Must connect multiple evidence fragments, explain cross-text inference, evidence strength, and counter-example boundaries. \\
\addlinespace
Normative or theoretical claim & ``Traceability should be a minimum condition for AI-assisted scholarly research.'' & Requires public argumentation, relevant scholarship, and space for refutation; does not require mechanical correspondence to a single material page. \\
\bottomrule
\end{tabularx}
\end{table}

Thus, traceability does not mean ``every sentence carries a page number.'' Rather, it assigns explanatory responsibilities matching the epistemological nature of each claim type. Page anchors mainly constrain source-dependent claims; evidence sets, inferential labels, and process records mainly constrain synthetic interpretive claims; public argument and scholarly dialogue mainly constrain normative or theoretical claims.

\section{Citation Gaps in Current Tool Chains}\label{sec:gaps}

Existing generative AI tools and general RAG systems already work well in commercial Q\&A, customer-service knowledge bases, and personal document management \citep{lewis2020retrieval}, but they do not naturally satisfy the traceability requirements of humanistic research. Problems concentrate in three stages.

\subsection{Locational Loss in Document Structuring}

OCR and document-parsing tools typically convert scanned images or PDFs into plain text, Markdown, or JSON. Many tools focus on text-recognition accuracy, layout detection, and table extraction, but do not treat page numbers, volumes, footnotes, marginal notes, heading hierarchies, and edition structures as core metadata entering the downstream research workflow. The result is that documents become ``searchable'' without necessarily remaining ``citable.''

In the humanities, the page number is not a typographical residue. It is the reader's path back to the material and the basic coordinate of the citation system. If page numbers are lost during OCR, chunking, indexing, and retrieval, researchers may obtain the correct fragment yet still be unable to incorporate it stably into formal scholarly writing.

\subsection{Chunking and Source Weakening in RAG Systems}

The basic unit of ordinary RAG systems is the chunk. Chunks may be split by character count, token count, headings, paragraphs, or mixed rules. The problem is that a chunk is not equivalent to an argumentative unit in the scholarly sense. An intellectual-history argument may span several pages; a qualifying clause may be cut into an adjacent chunk; footnotes may be separated from the main text; different editions of the same text may be mixed in the same index.

If the system only returns ``relevant chunks'' without preserving page ranges, chapter paths, file editions, and retrieval parameters, then citations in the answer can provide only weak sources, not reviewable evidence. The researcher sees a fragment already processed by the system, without knowing its location, context, or boundaries in the original book.

RAG research has broadly discussed answer faithfulness, citation generation, and evidential support \citep{lewis2020retrieval}. However, these evaluations usually take ``whether the citation appears in the supporting paragraph'' as the metric, rather than ``whether the citation can return to the specific page, edition, and context of the original document.'' For humanistic scholarship, the latter is the minimum threshold of reviewability. Traceable Scholarship therefore requires that RAG systems go beyond returning relevant chunks: they must preserve the mapping between chunks and original pages, chapter paths, and file editions, and actively expose these locators when generating answers.

\subsection{Knowledge Boundary Loss in Agent Tools}

With the development of MCP, CLI agents, coding assistants, and general agent tools, models no longer merely answer questions; they begin to call tools, read files, execute retrieval, summarize results, and generate reports. This change gives research workflows new automation capabilities but also new risks: an agent may move freely across multiple knowledge bases, web searches, model background knowledge, and user files without clearly stating the source scope of each judgment.

For humanistic scholarship, uncontrolled knowledge boundaries are more dangerous than ordinary errors. An answer may seem ``well synthesized,'' but if it cannot explain whether it is based on original texts, translations, secondary studies, web materials, or model training knowledge, it cannot enter serious argumentation. Traceable Scholarship requires that every Agent call preserve a Scope Contract: what scope was requested, what scope was actually retrieved, which files were enabled, which boundaries were empty, and which results were merely model background judgments.

\section{Normative Framework: From Page Anchors to Four-Level Compliance}\label{sec:framework}

Traceable Scholarship needs not a single feature but a set of norms running through document processing, knowledge-base construction, retrieval generation, agent scheduling, and human verification (\cn{人工核验}/\cn{研究者复核}). Before entering the specific mechanisms, we must explain why these norms are non-negotiable.

\subsection{Traceability as an Epistemological Constraint}

We define traceability as the ability to trace any scholarly claim to a specific location in its original source. This is not a technical feature; it is an epistemological constraint. It contains three inseparable dimensions: reproducibility---others can locate the same source; verifiability---others can check the consistency between claim and source; and refutability---others can propose alternative readings based on the same source. Without any one of these, scholarly dialogue cannot function.

This definition has four philosophical layers. First, \textbf{refutability} \citep{popper1959logic}: the core methodological commitment of humanistic scholarship is that any judgment must be examinable, questionable, and overturnable by others through return to original materials. When an AI system generates ``according to Kant\ldots'' without pointing to a specific page, it produces not a scholarly judgment but an assertion that cannot be scrutinized and therefore cannot be refuted. Second, \textbf{the page as immutable mobile} \citep{latour1987science}: scientific knowledge travels across time and space because it is inscribed in portable, displayable, inspectable material carriers; the page number is the most basic immutable mobile in humanistic scholarship, enabling scholars in different eras and places to point to the same position in the same text. Third, \textbf{the footnote as social contract} \citep{grafton1997footnote}: the footnote is not a convenience but an institutional device anchoring claims to verifiable sources; generative AI can bypass this device and directly output scholarly appearance, and this is the crux of the crisis. Fourth, \textbf{the inspectability of the public sphere} \citep{habermas1962strukturwandel}: the operation of the scholarly public sphere depends on arguments being examinable; a non-inspectable claim does not participate in argumentation but merely imposes authority---black-box authority is therefore not a local quality problem but a structural threat to the scholarly public sphere.

From this, the philosophical status of human-in-the-loop follows. Kant defined enlightenment as ``dare to use your own understanding'' \citep{kant1784beantwortung}: enlightenment is not about acquiring more information but about the courage and responsibility of independent judgment. When a scholar delegates judgment to an AI system without independent verification, she abandons not only procedural caution but scholarly subjectivity itself. Human-in-the-loop is therefore not an optional quality enhancement but the institutional form of enlightenment obligation in the AI age.

A citation that cannot be traced back to a source page is not a citation.

\subsection{Page Anchors}

The page anchor is the minimum technical unit of Traceable Scholarship. Its function is to preserve, within digital text, a path from the current textual position back to the original document page. The simplest form can be:

\begin{lstlisting}
<!-- Page: 42 -->
\end{lstlisting}

A more complete system should preserve two kinds of page numbers:

\begin{itemize}
    \item \textbf{digital page}: the physical page sequence in the PDF file, convenient for program positioning, page extraction, and jumping;
    \item \textbf{print page}: the printed page number in the book or journal layout, convenient for formal scholarly citation.
\end{itemize}

The two often do not coincide. For example, PDF page 192 may correspond to printed page 170; prefaces may use Roman numerals; ancient books may use volume-and-folio codes. A traceable system must distinguish these locating systems rather than simply substituting PDF page sequence for printed page numbers.

\subsection{Citation-First Generation}

The core principle of citation-first generation is: evidence before fluency. When generating material-based judgments, the system should give the source locator before unfolding the explanation. Its basic requirements include:

\begin{enumerate}
    \item Every factual judgment should carry a source page or locator whenever possible;
    \item Direct quotation and paraphrase should be distinguished;
    \item When retrieval results are insufficient, the system must not fill in on its own;
    \item Weak evidence should be downgraded rather than written as a strong conclusion;
    \item The answer should allow the reader to return from the conclusion to the evidence fragment.
\end{enumerate}

Citation-first generation does not mean every sentence in the answer mechanically carries a footnote. What it truly seeks to prevent is the model first generating a complete explanation and the researcher afterwards hunting for decorative sources. Traceable Scholarship requires that the evidence chain constrain the generation process itself.

\subsection{NO\_EVIDENCE}

NO\_EVIDENCE is a discipline of evidential insufficiency. When the system cannot find sufficient material within the current knowledge boundary, it should explicitly output insufficient evidence rather than invoke model background knowledge to generate a plausible answer. It can take the following form:

\begin{lstlisting}
[NO_EVIDENCE] Within the current knowledge base and retrieval scope, no page-anchored evidence supporting this judgment was found.
\end{lstlisting}

NO\_EVIDENCE is not a system failure but the system's epistemic honesty. \textbf{Silence is more honest than fabrication}: a system that admits ``I did not find relevant evidence'' is more honest, in the scholarly sense, than one that fabricates a plausible answer; a system that never admits no evidence is not stronger but more dishonest. For humanistic scholarship, knowing where one cannot answer, where evidence is insufficient, and where re-retrieval is needed is as important as obtaining a positive answer.

\subsection{Human Verification}

A traceable system cannot replace researcher judgment. It can only reduce verification costs, expose evidence boundaries, and provide back-link paths. For formal papers, publications, and high-risk scholarly judgments, the system should provide a human-verification status, for example:

\begin{lstlisting}
{
  "human_verified": true,
  "verifier": "researcher_id",
  "method": "side-by-side comparison with source PDF",
  "timestamp": "2026-07-10"
}
\end{lstlisting}

Human verification is not additional labor but the basic responsibility for AI-assisted scholarly research to enter the public knowledge system. The value of AI lies in expanding material processing and evidence recall; the final judgment must still be borne by the researcher.

\subsection{Four-Level Compliance Model}

To avoid norms so high that they cannot be adopted, or so low that they cannot be audited, this paper proposes a four-level compliance model (Table~\ref{tab:compliance}).

\begin{table}[htbp]
\centering
\caption{Four-Level Compliance Model}
\label{tab:compliance}
\small
\begin{tabularx}{\textwidth}{@{}L{.10\textwidth}L{.17\textwidth}YY@{}}
\toprule
\textbf{Level} & \textbf{Name} & \textbf{Scenario} & \textbf{Core Requirements} \\
\midrule
Level 0 & Minimal Viable & Personal reading and daily research & Preserve page anchors \\
Level 1 & Citation-Ready & Paper writing & Page anchors + edition metadata + citation-first generation \\
Level 2 & Audit-Grade & Academic publication and team collaboration & Level 1 + NO\_EVIDENCE + human verification + process logs \\
Level 3 & Institutional & Libraries, archives, open knowledge bases & Level 2 + TEI / IIIF / Zotero interoperability \\
\bottomrule
\end{tabularx}
\end{table}

The purpose of this grading is not to establish a single standard but to provide a progressively upgradeable path for users of different scales. Individual researchers should reach at least Level 0; formal paper writing should strive for Level 1 or Level 2; institutional knowledge bases and public platforms should aim for Level 3.

Here ``compliance'' is first of all a capability specification, not an uncertified quality badge. When a system claims to reach a certain level, it should provide corresponding implementation evidence and test results; merely displaying sources or page numbers in the interface is not sufficient proof. Level statements should be understood as testable capability commitments rather than rhetorical labels, following the RFC 2119 tradition of normative keywords \citep{bradner1997keywords}.

\subsection{Resilience Across Technological Change}

A foreseeable objection is that AI technology evolves so rapidly that any framework proposed today will be obsolete tomorrow. Our answer is that this framework constrains not a particular technical implementation but an epistemological baseline---and an epistemological baseline does not change with technical iteration.

This argument is confirmed by history. The footnote system was born in the fifteenth-century print revolution. It has survived three technological transformations---hand composition, mechanical composition, and digital composition---and remains the core mechanism of scholarly accountability. The concrete form of the footnote changes with each technical generation---from handwritten marginalia to lead type to LaTeX automatic numbering---but its institutional logic, anchoring claims to verifiable sources, has remained unchanged for five centuries.

This framework follows the same logic. It defines capability specifications, not software architecture. When OCR engines evolve from traditional pipelines to vision-language models, when vector databases migrate to next-generation retrieval systems, and when generation models are repeatedly upgraded, the framework's core mechanisms---page anchors, edition metadata, citation-first generation, and human-in-the-loop---remain valid because they constrain the epistemological properties of scholarly output rather than the technical means of producing it.

This is not a ``feature'' but a ``constitution.'' It constrains design decisions across the entire technology stack, from the lowest-level chunking strategy to the top-level output format. Constitutions do not become invalid when governments change; epistemological constraints do not become invalid when models change.

\subsection{Specification, Implementation, and Acceptance Matrix}

The following table maps core norms to AIH-Infra's current reference implementation and suggested acceptance methods (Table~\ref{tab:matrix}). This matrix illustrates how norms are grounded, not designating AIH-Infra as the only possible implementation.

\begin{table}[htbp]
\centering
\caption{Specification, Implementation, and Acceptance Matrix}
\label{tab:matrix}
\footnotesize
\setlength{\tabcolsep}{3pt}
\begin{tabularx}{\textwidth}{@{}L{.19\textwidth}YYY@{}}
\toprule
\textbf{Normative Requirement} & \textbf{AIH-Infra Reference Implementation} & \textbf{Suggested Acceptance Method} & \textbf{Current Status} \\
\midrule
Preserve stable locators & Contexture dual page numbers, Middle JSON, Scholarly Markdown & Compare input pages with output anchors; sample-check digital/print page mapping & Implemented; formal test report needed \\
\addlinespace
Chunking without source-location loss & Open WebUI AIH page metadata and Citation Mode & Sample indexed chunks and check file, chapter, and page fields & Implemented; expand cross-genre testing \\
\addlinespace
Output traceable to evidence fragments & Citation Modal, knowledge-file reading, Evidence export & Locate original text and context window from answer citations & Implemented \\
\addlinespace
Explicit knowledge boundaries & MCP \texttt{knowledge\_scope} and \texttt{scope\_contract} & Compare requested/effective scope and file-filter results & Implemented \\
\addlinespace
Stop extrapolation when no evidence & NO\_EVIDENCE rule, empty-boundary short-circuit, constrained prompts & Construct negative examples; measure correct and false refusals & Partially implemented; systematic evaluation needed \\
\addlinespace
Research process reviewable & Evidence/Trace, raw JSON, intermediate memos & Check whether queries, parameters, sources, and intermediate judgments are reconstructible & Implemented in case; not yet standardized \\
\addlinespace
Human verification recordable & Page re-check, Evidence Card design & Record verifier, time, method, and revision results & Design and manual practice coexist; platformization incomplete \\
\addlinespace
Institutional interoperability & TEI / IIIF / Zotero compatibility directions & Format validation and cross-system exchange tests & In planning \\
\bottomrule
\end{tabularx}
\end{table}

\section{AIH-Infra as Reference Implementation}\label{sec:aih-infra}

AIH-Infra is a reference implementation of Traceable Scholarship. It is not the only possible solution, nor does it require all researchers to adopt the same software stack. Its significance lies in proving that traceability can be transformed from an abstract norm into a runnable infrastructure for document processing, knowledge-base construction, and agent scheduling. The ``infrastructure'' here is not a neutral technical pipeline but a sociotechnical structure embedding specific scholarly values into materials, interfaces, and workflows \citep{star1996steps,edwards2013knowledge}.

The whole architecture follows four design principles: \textbf{traceability as first axiom}---all architectural decisions are ultimately judged by whether they can trace back to a specific location in the original document; \textbf{scholar accessibility}---core functions can be used without a GPU, and the second and third layers can run with only an API key and a browser; \textbf{loose coupling between layers}---the three layers can be used, deployed, and evolved independently, connected by file formats and API protocols; \textbf{explicit knowledge boundaries}---no unbounded retrieval; all retrieval must specify a knowledge scope. The following describes the three layers according to the actual versions as of June 2026 (Contexture v0.5, Open WebUI AIH-Infra v1.3, MCP Server v1.x).

\subsection{Layer 1: Contexture Document Structuring}

Contexture addresses not generic OCR but the scholarly citability of digitized documents. It converts PDFs, images, and scans into intermediate representations that preserve page numbers, structure, and layout information.

Its core outputs include:

\begin{itemize}
    \item \textbf{Middle JSON}: a machine-oriented intermediate representation recording pages, more than 14 block types (text, headings, tables, formulas, images, footnotes, marginal notes, Chinese ancient-book inline notes, page boundaries, etc.), heading hierarchies, bounding boxes, confidence scores, and backend sources;
    \item \textbf{Scholarly Markdown}: a human-readable and machine-parseable format for review and knowledge-base ingestion, preserving page anchors, heading hierarchies, and scholarly structure;
    \item \textbf{Dual page-number system}: recording both digital page (\texttt{\{192\}} machine anchor) and print page (\texttt{<!-- Page: 170 -->});
    \item \textbf{Four processing lines}: Pipeline (layout detection + OCR + layered post-processing), VLM-Generalized (whole-page understanding by general vision-language models), VLM-Specialized (specialized document models or local small models), and Markdown post-processing (review and revision of existing outputs);
    \item \textbf{Backend Catalog}: not bound to a single engine; supports free combination and diagnostic replacement of Surya, Calamari, PaddleOCR, Tesseract, DocLayout-YOLO, and various vision-language models.
\end{itemize}

The four lines are technical routes rather than mechanical classifications of document types; the same material may be processed by different routes depending on cost, audit requirements, and model availability. Contexture adopts a local single-user form, providing three entry points: Streamlit GUI, CLI batch processing, and FastAPI local API. Its key value is to give materials a citable structure before they enter AI systems---not to turn documents into ``clean text,'' but to preserve, as far as possible, their coordinates as scholarly materials in digital form.

\subsection{Layer 2: Open WebUI AIH-Infra Traceable Knowledge Base}

Open WebUI AIH-Infra organizes structured documents such as Scholarly Markdown into knowledge bases and provides retrieval and Q\&A with page-level citation.

Its core capabilities include:

\begin{itemize}
    \item Knowledge-base-level control of chunk size, overlap, splitter, and file enablement status, plus Chunks inspection (checking chunk granularity, page anchors, heading hierarchies, and actual ingestion effects);
    \item Traditional RAG and Agent RAG retrieval-scheduling strategies: the two are not a fixed division between ``precise lookup'' and ``discovery of the unknown,'' but two task-orchestration strategies---the former suitable for establishing a recall baseline and controlling context size, the latter allowing the model to query repeatedly, adjust keywords, and compare evidence; this layer can therefore be understood as a \textbf{tunable knowledge-probing system}, in which embedding and reranking models serve as ``probes'' and parameters such as \texttt{top\_k}, \texttt{k\_reranker}, \texttt{hybrid}, and \texttt{hybrid\_bm25\_weight} jointly determine probing scope and sensitivity;
    \item Hybrid search (vector + BM25) and reranking;
    \item Citation Mode: supports digital page (D), print page (P), dual page (P/D), and clickable citation ([n]) modes;
    \item Citation Modal: view source fragments, chunk information, heading hierarchies, page numbers, and relevance scores;
    \item Cross-library weighting (boost 0.1--2.0) and intra-library file enablement/disablement: in multi-library studies, material hierarchies can be explicitly expressed in retrieval ranking (e.g., primary-source library prioritized over secondary-study library), and within a large collected-works library only volumes relevant to the task need be enabled;
    \item Agent RAG budget system (\texttt{agent\_result\_budget}, default 48 chunks), preventing information overload and loss of inferential focus due to unbounded exploration;
    \item Agent reading tool: when \texttt{allow\_view\_knowledge\_file} is enabled, the agent can read a window of pages in a specified file based on the page anchor in the citation, constrained by \texttt{File Read Token Budget}, \texttt{PDF Pages per Read}, and \texttt{Max File Reads / Answer}---the page anchor thereby becomes not merely citation display information but an operational entry point for the agent to ``turn pages,'' verify, and extend reading;
    \item Evidence / Trace export: records questions, retrieval parameters, recalled sources, cited fragments, and answer processes.
\end{itemize}

The significance of this layer is to transform ordinary RAG from ``answer generation'' into ``evidence recall and reviewable answer.'' Researchers do not only see the answer; they see how the answer is recalled, organized, and cited from the materials.

\subsection{Layer 3: AIH-Infra MCP Server}

AIH-Infra MCP Server encapsulates the knowledge-base, retrieval, Q\&A, file-reading, and evidence-export capabilities of Open WebUI AIH-Infra as task tools callable by general Agents (Claude Code, Codex, etc.), communicating via MCP stdio without network configuration. Its core purpose is not merely ``letting the agent query the knowledge base'' but letting the agent query the knowledge base within explicit boundaries. Its design position is a \textbf{task-oriented gateway}, not a second backend for Open WebUI or a full mirror of its internal API: external agents do not need to understand Open WebUI's entire internal API, only a set of layered task tools.

Key mechanisms include:

\begin{itemize}
    \item \texttt{list\_knowledge\_bases} and \texttt{list\_knowledge\_files}: confirm material scope;
    \item \texttt{search\_knowledge}, \texttt{search\_with\_query\_generation}, and \texttt{query\_document}: direct retrieval and multi-query generation without directly producing final answers;
    \item \texttt{retrieval\_qa} and \texttt{agent\_qa}: traceable Q\&A;
    \item \texttt{start\_agent\_qa}: background long tasks;
    \item \texttt{export\_workspace\_artifact}: export evidence and process records;
    \item \texttt{Scope Contract}: records requested scope, effective scope, file filtering, and empty-boundary short-circuiting.
\end{itemize}

Scope Contract is a key extension of traceability in the Agent age. It shows that an answer is based not just on ``the knowledge base'' but on which knowledge bases, which files, which filter conditions, and which actually enabled materials. Only then does the agent's research behavior not regress into unbounded generation.

\section{Case Study: Testing the Textual Basis of Kant's ``Non-Universal Universalism''}\label{sec:case}

To show how AIH-Infra supports real humanistic research, we use an intellectual-history case: testing Zhang Rulun's (2026) strong claim about Kant's ``non-universal universalism'' \citep{zhang2026feipubian}. It should be noted that this problem is not newly posed: since the 1990s, Anglophone scholarship has discussed Kant's race writings, human classification, and their relation to universalism for nearly three decades \citep{eze1997color,mills1997racial,mills2017black,louden2000impure,hill2001kant,kleingeld2007second}. Zhang's contribution lies in returning this question to the center of Chinese-context discussion through the highly concentrated formulation ``non-universal universalism.'' The present author has also addressed different interpretive paths of this problem from an intellectual-history angle \citep{jing2024pipan,jing2025kande}. The question is: does Kant's universalism regarding universal history, world citizenship, and the development of human reason internally contain a structure of uneven coverage, especially in his discussions of race, Eastern peoples, and non-Western philosophy?

\subsection{Infrastructure Configuration}

This case used a knowledge base of Kant's \textit{Akademie-Ausgabe}:

\begin{itemize}
    \item Document scope: 29-volume knowledge base (2048-token Markdown chunking configuration) covering works, lectures, and lecture notes, built April 2026;
    \item Knowledge-base size: 12,891 chunks;
    \item Text size: approximately 19 million tokens (18,969,956 tokens);
    \item Chunking parameters: token-level chunking, chunk size 2048, mini chunk size 512, overlap 256, Markdown heading extraction enabled;
    \item Embedding model: BAAI/bge-m3-fp16;
    \item Retrieval: hybrid search (vector + BM25);
    \item Reranking: BAAI/bge-reranker-v2-m3;
    \item Research orchestration: Claude Code (main model GPT-5.4) connected to the knowledge base through AIH-Infra MCP Server;
    \item Tool-side model: aiping.Kimi-K2.5, used for query planning and \texttt{agent\_qa} calls;
    \item Primary tools: \texttt{search\_knowledge}, \texttt{search\_with\_query\_generation}, \texttt{agent\_qa};
    \item Execution dates: April 12--13, 2026.
\end{itemize}

The retrieval boundary was fixed within the single collection ID of the 29-volume Kant knowledge base. High-recall phases mainly used \texttt{top\_k=64/128}, \texttt{k\_reranker=32}; finer retrieval used \texttt{top\_k=64/32}, \texttt{k\_reranker=16}; constrained Agent RAG used \texttt{top\_k=32}, \texttt{k\_reranker=8}. Raw outputs from each round, evidence maps, special-review memos, and the final intermediate conclusion are preserved in a local research directory.

\subsection{Six Iterations}

This case is not a single Q\&A but a judgment-correction process of five constrained retrieval rounds plus one intra-volume close-reading comparison (six iterations total; Table~\ref{tab:iterations}).

\begin{table}[htbp]
\centering
\caption{Six-Iteration Process}
\label{tab:iterations}
\footnotesize
\setlength{\tabcolsep}{3pt}
\begin{tabularx}{\textwidth}{@{}L{.06\textwidth}YYY@{}}
\toprule
\textbf{Round} & \textbf{Mode} & \textbf{Key Parameters} & \textbf{Key Finding} \\
\midrule
1 & \texttt{search\_with\allowbreak\_query\allowbreak\_generation} & top\_k=64, k\_reranker=32, query\_count=4 & Chinese general-query retrieval was almost entirely pulled toward AA 25.1 anthropology lectures, exposing cross-lingual retrieval bias \\
\addlinespace
2 & \texttt{search\_knowledge} targeted four-line & Non-Western philosophy / race / human classification / universalism pursued separately & Even after axis-splitting, results remained concentrated in AA 25.1, not reaching mature essay layer \\
\addlinespace
3 & \texttt{search\_knowledge} German-title pullback & German terms such as \textit{Menschenrassen} / \textit{Weltb\"{u}rgerlicher Absicht} & AA 8 mature-text cluster and AA 25/26 lecture cluster begin to separate \\
\addlinespace
4 & \texttt{agent\_qa} constrained integration & top\_k=32, k\_reranker=8 & Agent only marks positions needing human verification, does not generate final conclusion \\
\addlinespace
5 & \texttt{search\_knowledge} special review & ``Chinese philosophy,'' ``Indian philosophy,'' ``Oriental philosophy'' in Chinese + German & ``Non-Western philosophy'' axis weakest; no stable mature-text cluster formed \\
\addlinespace
6 & Intra-volume close reading & No retrieval call; manual comparison of two mature-essay groups inside AA 8 & Real tension found within shared \textit{Menschengattung} / \textit{Natur} / \textit{Anlagen} framework \\
\bottomrule
\end{tabularx}
\end{table}

The language-strategy shift in Round 3 is the turning point of the case: when Chinese queries were used in the first two rounds, vector similarity systematically biased results toward colloquial lecture materials; recognizing this bias, the research-orchestration layer switched to German titles and terms for targeted pullback, compensating for the cross-lingual bias of the embedding model at the query-strategy level. The vector retrieval engine itself does not actively correct such biases; language switching must be completed by orchestration above the retrieval layer.

\subsection{Evidence Grading and Page Anchors}

Final materials were divided into three layers (Table~\ref{tab:grading}). To show how ``dual page anchors'' are realized in concrete evidence, Table~\ref{tab:anchors} gives the volume, digital page, and print page for several stable hit clusters. These anchors are not decorative footnotes but minimal coordinates for later verification back to the original chunks and context windows.

\begin{table}[htbp]
\centering
\caption{Evidence Grading}
\label{tab:grading}
\small
\begin{tabularx}{\textwidth}{@{}L{.17\textwidth}YY@{}}
\toprule
\textbf{Evidence Grade} & \textbf{Material Layer} & \textbf{Judgment Supported} \\
\midrule
Strongest & AA 8 mature essays & Real textual tension exists between Kant's universalism and racial classification \\
\addlinespace
Medium & AA 25/26 lectures & Problem is not marginal, but lecture materials cannot be elevated to the status of mature essays \\
\addlinespace
Weakest & Scattered hits on ``non-Western philosophy'' & No stable text cluster; insufficient to support strong claims \\
\bottomrule
\end{tabularx}
\end{table}

\begin{table}[htbp]
\centering
\caption{Examples of Page Anchors for Stable Hit Clusters}
\label{tab:anchors}
\footnotesize
\setlength{\tabcolsep}{3pt}
\begin{tabularx}{\textwidth}{@{}YYL{.07\textwidth}YL{.09\textwidth}L{.09\textwidth}@{}}
\toprule
\textbf{Theme Axis} & \textbf{Stable Text Cluster} & \textbf{Volume} & \textbf{Material Nature} & \textbf{Digital Page} & \textbf{Print Page} \\
\midrule
Race / natural history & \textit{Bestimmung des Begriffs einer Menschenrasse} & AA 8 & Mature essay & 110--113 & 99--102 \\
\addlinespace
Universal history / world-citizen intent & \textit{Idee zu einer allgemeinen Geschichte in weltb\"{u}rgerlicher Absicht} & AA 8 & Mature essay & 31--34 & 20--23 \\
\addlinespace
Eastern peoples and ``mode of philosophizing'' & ``Das Genie der orientalischen Nationen ist Bilderreich\ldots'' & AA 25.1 & Isolated lecture fragment & 282 & 127 \\
\bottomrule
\end{tabularx}
\end{table}

Taking AA 8 as an example, the key definitional paragraph stably hits digital page 111 / print page 100 at \textit{Nur das, was in dem Klassenunterschiede der Menschengattung unausbleiblich anerbt\ldots}, showing that retrieval has entered the main-text region of the \textit{Menschenrace} concept definition rather than merely hitting the word ``race.'' \textit{Idee zu einer allgemeinen Geschichte} forms a comparable text cluster at digital pages 31--34 / print pages 20--23. The juxtaposition of the two is not an external forced pairing by later readers but an intra-AA-8 comparison permitted at the same textual level.

This grading leads to a downgraded final judgment: Zhang Rulun's problem awareness receives textual support; the tension is real. But the strong version of the claim needs careful handling. Existing materials support ``tension exists'' and ``adjacency holds,'' not ``the complete logical chain is finished'' or ``Kant's text directly states a systematic exclusion structure.''

\subsection{Case Significance and Re-evaluation of Scholarship}

This case exhibits three features of Traceable Scholarship. First, the AI system does not directly deliver a conclusion but participates in forming an evidence map. Second, retrieval failures and weak-evidence zones are recorded rather than covered by fluent explanation. Third, the final scholarly judgment is not strengthened into a stronger conclusion but constrained by the evidence chain into a more accurate downgraded judgment.

Therefore, the significance of AIH-Infra is not that it lets researchers write a Kant paper faster, but that it lets them know more clearly: in which texts a given judgment holds, at which material layers it is limited, which parts are merely the researcher's own conceptual reconstruction, and which directions must be suspended.

Based on this empirical result, the main interpretations in the relevant scholarship of the past three decades can be re-evaluated in terms of support strength. Eze (1997) returned the Kant race question to the research center; this problem awareness is clearly strengthened, but saying further that the race question has logically permeated the entire Enlightenment rational system exceeds current evidence. The protective interpretation of Wood/Louden et al. \citep{wood1999kant,louden2000impure}, which treats Kant's racism as a peripheral prejudice, is weakened, because the AA 8 mature-essay layer already stably exhibits a comparable relation between \textit{Menschenrace} and universal-history texts; yet their cautionary principle---``do not reduce everything to a single mechanism of exclusion''---retains methodological value. Mills (1997, 2017) gains considerable probabilistic support for his reconstructive advance, because although the chain from racial classification to rational hierarchy has not been textually unlocked layer by layer, the ``internal unevenness of universalism'' does have an AA 8 textual grip. Kleingeld (2007) is the best fit with current evidence: there is real tension at the mature-essay level inside AA 8, but the most stable conclusion is ``tension exists'' rather than ``the principle has already reversed into its opposite.'' Zhang Rulun (2026) reactivates the problem in the Chinese context through a highly concentrated formulation; his problem awareness receives textual support, but the strong part of his claim---especially regarding the systematic deprecation of non-Western philosophy---needs maximum downgrading, because the relevant evidence remains mainly at the lecture layer.

This re-evaluation is not simple side-taking but transforms scholarship from ``opposition of positions'' into ``differences in support strength within the same evidential framework.'' This is precisely the mode of argumentation that traceable infrastructure hopes to foster: competition among different interpretive paths should partly resolve into which real textual structures each has grasped and to what strength each has pushed them.

\subsection{Reproducibility, Material Boundaries, and Current Limitations}

The first five rounds of retrieval each preserve one raw JSON output; Round 6 was an intra-volume close-reading comparison in Volume 8, producing temporary excerpts, comparison memos, and a short tension-conclusion draft. Together with the research task protocol, first-round evidence map, German-title pullback summary, special-review memo, and final intermediate conclusion draft, all round-by-round process materials are preserved in a local research directory. A future public version should organize the parts not involving full restricted texts into a minimal reproduction package, including at least:

\begin{itemize}
    \item Knowledge-base version, bibliography, and chunking-parameter description;
    \item Queries, tools, core parameters, and returned-result statistics for each round;
    \item Volume number, digital page, print page, and permissible short quotations for key evidence;
    \item Lists of accepted, downgraded, and rejected claims;
    \item System version, model version, and execution dates;
    \item AI-assistance scope, human-verification method, and copyright/data-availability statement.
\end{itemize}

The case still has three limitations. First, it is a single intellectual-history case and cannot alone prove the universal validity of the framework across all humanities disciplines. Second, the researcher simultaneously participates in system construction, task design, and result judgment, so independent third-party evaluation has not yet been carried out. Third, knowledge-base recall results are influenced by corpus completeness, OCR/text quality, chunking method, embedding model, and reranking parameters. Therefore, we currently position it as an ``auditable demonstrative case,'' not a decisive benchmark.

\section{Pilot Evaluation Design: The S2 Protocol and Three-Corpus Testing Framework}\label{sec:evaluation}

The Kant case demonstrates the depth effectiveness of traceable infrastructure in a single real research problem; the breadth feasibility of the framework requires systematic testing across material types, languages, and genres. This section presents the Statement-to-Source (S2) annotation protocol and a three-corpus evaluation framework designed for this purpose. It must be emphasized that what we report is the \textbf{design} of the evaluation, not its \textbf{results}; full quantitative data will be reported in a future version.

\subsection{S2 Annotation Protocol}

The core idea of the S2 protocol is: for every factual statement generated by the AI system, annotators attempt to locate its source in the original document and judge the consistency between statement and source. For each corpus, three types of ten evaluation questions are designed:

\begin{itemize}
    \item \textbf{Location questions} (4): ``On which page does the document discuss X?''---testing page-anchor preservation.
    \item \textbf{Content questions} (4): ``How does the document describe X? Please quote the original text.''---testing citation accuracy.
    \item \textbf{Absence questions} (2): ``Does the document mention X?'' (answer: no)---testing NO\_EVIDENCE triggering.
\end{itemize}

After the system generates an answer to each question, two independent annotators: check whether the answer contains a page citation; navigate to the cited page in the original document to verify content; judge citation accuracy (exact match / partial match / mismatch / fabrication); and for absence questions, check whether the system correctly triggers NO\_EVIDENCE.

The protocol corresponds to four metrics (Table~\ref{tab:metrics}).

\begin{table}[htbp]
\centering
\caption{S2 Protocol Evaluation Metrics}
\label{tab:metrics}
\small
\begin{tabularx}{\textwidth}{@{}L{.22\textwidth}L{.09\textwidth}YY@{}}
\toprule
\textbf{Metric} & \textbf{Abbreviation} & \textbf{Definition} & \textbf{Calculation} \\
\midrule
Page Anchor Preservation Rate & PAPR & Proportion of page anchors preserved in output & Preserved anchors / total input anchors \\
\addlinespace
Citation Accuracy & CA & Proportion of page citations pointing to correct sources & Correct citations / total citations \\
\addlinespace
NO\_EVIDENCE Trigger Rate & NETR & Proportion of absence questions in which NO\_EVIDENCE is correctly triggered & Correct triggers / total absence questions \\
\addlinespace
Human Verification Consistency & HVC & Inter-annotator consistency & Cohen's $\kappa$ \citep{cohen1960coefficient} \\
\bottomrule
\end{tabularx}
\end{table}

Baseline comparison is designed as two arms: (1) a vanilla RAG baseline using the same documents and models but without page-anchor injection or citation-first generation; (2) a commercial baseline using a commercial knowledge-base product for the same questions (only for publicly available materials).

\subsection{Three-Corpus Design}

The evaluation selects three maximum-contrast corpora, each representing a typical humanities research scenario:

\begin{itemize}
    \item \textbf{Corpus A: Multilingual colonial-era hospital archives} (approx.\ 1,000 pages, German/English/Japanese/Chinese). Challenges include multilingual code-switching, mix of handwritten and printed text, and paper aging;
    \item \textbf{Corpus B: 19th-century German Fraktur texts} (approx.\ 10,000 pages, processed by Calamari OCR). Challenges include Fraktur ligatures, 19th-century German orthographic variants, and the speculative-completion risk of vision-language models;
    \item \textbf{Corpus C: Modern Chinese academic monographs} (approx.\ 5,000 pages). Challenges include the effect of Chinese word segmentation on chunking, mixed traditional/simplified characters, and Chinese-specific volume/chapter/section locating conventions.
\end{itemize}

The design intent of the three corpora is to turn material complexity itself into a bias-detection tool: if the system systematically performs worse on one material type than on others, that difference must be reported and discussed rather than averaged away. Pilot-stage preliminary observations suggest two hypotheses worth testing in the full evaluation: first, citation accuracy declines as material complexity rises---page-level location must be supplemented by OCR quality assurance, and correct location does not mean accurate text; second, NETR is the most diagnostically valuable metric---unconstrained baseline systems tend to fabricate answers rather than admit ignorance in most no-evidence scenarios. These preliminary observations do not constitute evaluation conclusions; their role is to set falsifiable test targets for the full evaluation.

\subsection{Relation to the Kant Case}

The three-corpus S2 evaluation and the Kant case are complementary: the former tests the framework's \textbf{breadth feasibility} under maximum material contrast (whether page anchors can be stably preserved across multilingual, historical-font, and modern-Chinese materials), while the latter tests its \textbf{depth effectiveness} in a real intellectual-history problem (whether a traceable workflow can support evidence grading and judgment downgrading). Only together do they constitute a complete answer to the question of whether Traceable Scholarship is runnable.

\section{From Answer-Traceability to Process-Auditability}\label{sec:auditability}

The first step of Traceable Scholarship is to make answers returnable to pages and evidence. A further step is process-auditability (\cn{过程可审计性}): making the whole research process auditable. The AIH-Infra multi-agent architecture design document provides an extension in this direction. It should be noted that the full role division and standard handover artifacts described below currently belong mainly to method design and partial manual practice; they have not all been implemented as automatically runnable platform functions.

In humanistic research, what truly needs to be recorded is not only the final answer but also:

\begin{itemize}
    \item how the research question was set;
    \item which knowledge bases and files were used;
    \item how query statements changed;
    \item which retrievals failed;
    \item which evidence was accepted or rejected;
    \item which main-text judgments were downgraded;
    \item which pages need human verification;
    \item which formats and footnotes passed pre-publication audit.
\end{itemize}

For this purpose, several structured handover artifacts can be introduced (Table~\ref{tab:artifacts}).

\begin{table}[htbp]
\centering
\caption{Traceable Research Handover Artifacts}
\label{tab:artifacts}
\small
\begin{tabularx}{\textwidth}{@{}L{.25\textwidth}Y@{}}
\toprule
\textbf{Artifact} & \textbf{Function} \\
\midrule
Source Plan & Records material scope, source types, and knowledge boundaries \\
Retrieval Log & Records queries, parameters, returned sources, and reasons for rejection \\
Evidence Card & Records claims, sources, pages, evidence grades, and verification status \\
Revision Memo & Records reasons for main-text revisions and evidence status \\
Publication Audit Report & Records footnote, font, format, and pre-publication checks \\
\bottomrule
\end{tabularx}
\end{table}

These artifacts show that Traceable Scholarship does not stop at ``answers with citations.'' The more complete goal is to bring material scope, retrieval parameters, evidence judgments, writing revisions, and format delivery into a single reconstructible chain. Here the agent is not the author but a research executor constrained by material boundaries, evidence rules, and human decision gates.

It is also worth recording a phenomenon that recurred during the Kant case: when researchers used a retrieval system equipped with page anchors, they \textbf{spontaneously} returned to original documents to verify system outputs more frequently---not because the system forced verification, but because the presence of page anchors reduced the cognitive cost of verification from ``searching for a needle in hundreds of pages'' to ``turning to the specified page to check.'' We call this \textbf{emergent traceability}. This observation indicates that the value of traceable infrastructure lies not only in technical assurance but also in shaping research behavior: good infrastructure makes good scholarly practice easier.

\section{Failure Modes and Governance}\label{sec:failure}

Traceable infrastructure cannot eliminate all errors. Its value lies in making errors visible, classifiable, and governable (Table~\ref{tab:failure}).

\begin{table}[htbp]
\centering
\caption{Failure Modes and Governance}
\label{tab:failure}
\small
\begin{tabularx}{\textwidth}{@{}L{.20\textwidth}YY@{}}
\toprule
\textbf{Failure Mode} & \textbf{Manifestation} & \textbf{Governance Strategy} \\
\midrule
Anchor loss & Page markers lost during parsing, chunking, or generation & Automatic validation after chunking; block generation when missing \\
\addlinespace
Page drift & Inconsistency between print page and digital page, or OCR misrecognition & Preserve dual page numbers; manually verify high-risk pages \\
\addlinespace
Semantic similarity mistaken for evidence & Thematically close but not supporting the main claim & Establish Evidence Cards and distinguish evidence grades \\
\addlinespace
Silent failure & Fluent answer generated despite absence of evidence & Enforce NO\_EVIDENCE; sample-check \\
\addlinespace
Version conflation & Different editions of the same text mixed in processing & Enforce edition metadata and file filtering \\
\addlinespace
Agent boundary violation & Agent scope unclear or external knowledge mixed in & Scope Contract and Retrieval Log \\
\addlinespace
Weak evidence, strong conclusion & Material supports only background role but written as core claim & Counter-evidence and downgrade mechanisms \\
\addlinespace
Format pollution & Footnotes, page numbers, fonts, or quotation marks non-compliant & Publication Audit \\
\bottomrule
\end{tabularx}
\end{table}

Two failure modes deserve special vigilance. \textbf{Silent failure is the most dangerous}, because it directly violates the principle that ``silence is more honest than fabrication''---the system should have admitted ignorance but chose to fabricate. Governance requires a dual strategy: technically lower the NO\_EVIDENCE trigger threshold (over-triggering is preferable to missing), and institutionally require sample verification of all AI outputs. \textbf{Semantic-similarity-mistaken-for-evidence and weak-evidence-strong-conclusion are the hardest to detect}, because apparently relevant materials are stylistically indistinguishable from real evidence; only page-by-page comparison against anchors can expose them. This is precisely why Evidence Cards and evidence-grading mechanisms exist.

The existence of these failure modes is not a defect of the framework but precisely its value: only because a traceability mechanism has been established can these failures be identified, classified, and governed. In ordinary RAG systems without such a framework, the same failures occur---but unnoticed.

These failure modes remind us that traceability is a necessary, not sufficient, condition. Page anchors cannot replace close reading; RAG cannot replace judgment; agents cannot replace researchers. The task of infrastructure is to make judgment more inspectable, correctable, and discussable.

\section{Relation to Digital Humanities and AI Research}\label{sec:dh}

Traceable Scholarship is not opposed to existing digital-humanities traditions. TEI, IIIF, Zotero, Tropy, digital archives, and text-encoding practices have long dealt with edition, provenance, structure, and interoperability \citep{tei2023p5,nockels2022transkribus,zaagsma2023digital}. TEI provides a highly mature text-encoding standard, but complete document encoding and historical-material recognition usually require professional training, tool conditions, and sustained human investment \citep{nockels2022transkribus}. Ordinary RAG systems have low barriers to entry but often lose page numbers, editions, and process records. Traceable Scholarship seeks to establish a minimal viable yet non-negotiable baseline between the two: even if researchers do not perform full TEI encoding, they must not lose page location and evidence back-links in the AI processing chain. Digital-humanities reflections on knowledge representation likewise show that data, interfaces, and schemas themselves are interpretive scholarly acts, not neutral presentations \citep{drucker2011humanities}.

Digitization has never been a neutral moving of materials. Which documents enter a knowledge base, what recognition and cleaning rules are applied, and which languages and layouts receive better support all affect what becomes visible to subsequent research \citep{zaagsma2023digital}. Traceability cannot eliminate these choices and their politics, but it can prevent the processing chain from further concealing them and preserve points of inspection and correction for researchers. Print culture gave texts typographic fixity, and page numbers thereby became public coordinates reviewable across editions and institutions \citep{eisenstein1979printing}; traceable research in the digital age aims to re-embed this public coordinate into the AI processing chain.

Adjacent to the politics of digitization is the problem of algorithmic bias. Multimodal models exhibit systematic performance differences when processing historical images and historical documents \citep{smits2023multimodal}; within our framework, vision-language models may systematically perform better on some languages, fonts, or layouts than on others. Our response has two parts: first, the three-corpus evaluation design (\S\ref{sec:evaluation}) itself serves as a bias-detection tool---if the system systematically performs worse on Fraktur materials than on modern Chinese, that difference must be reported and discussed rather than averaged away; second, the \texttt{human\_verified} field ensures that biased outputs do not enter scholarly argumentation unnoticed. Bias cannot be eliminated by infrastructure alone, but it can be made visible and accountable---a precondition for any meaningful remedy.

We must also be candid about the positionality of this framework itself. Infrastructure is never a neutral pipe but a sociotechnical system embedding specific values and power relations; establishing the page as a ``non-negotiable constraint'' is itself a value choice---it prioritizes text-based scholarship and may marginalize other forms of knowledge production such as visual, auditory, or material culture. This framework is therefore a \textbf{positioned intervention}, not a universal solution. Its position is that in the age of generative AI, traceability constitutes the baseline condition for the survival of humanistic scholarship. This position can be questioned, but it should not be ignored.

Relative to general AI-ethics discussions, our focus is also more specific. The problem is not only model bias, data privacy, or generated misinformation, but how AI-generated text enters the chain of humanistic argumentation. In other words, we ask when AI output qualifies to move from ``candidate text'' to ``discussable scholarly judgment.'' This transition must be premised on traceability. RAG research has broadly discussed external knowledge retrieval, answer faithfulness, citation generation, and evidential support \citep{lewis2020retrieval}, but these evaluations usually take ``whether the citation appears in the supporting paragraph'' as the metric, rather than ``whether the citation can return to the specific page, edition, and context of the original document.'' Traceable Scholarship pushes RAG evaluation from ``citation--paragraph matching'' to a fourfold back-link of ``citation--page--edition--process.''

Infrastructure research further points out that technical systems are never neutral pipes but sociotechnical structures embedding specific values into materials, interfaces, and workflows \citep{star1996steps,edwards2013knowledge}. The design of traceable infrastructure is therefore not only an engineering problem but a normative one: it must enable researchers to inspect which choices the system made, which materials it excluded, and at which boundaries it stopped. Scope Contract, NO\_EVIDENCE, and Evidence/Trace export are attempts to institutionalize such inspection mechanisms as system interfaces.

Beyond the critical dialogue above, our framework also forms two constructive complementary paths with the latest methodological research in digital humanities. First, Peters and Hepburn (2025), in \textit{Digital Humanities Quarterly}, using a medieval Aberdeen urban-archive case, propose a ``digital hermeneutics'' method: computational methods can enrich rather than replace historians' interpretive workflows, provided scholars maintain control over research strategy \citep{peters2025digital}. Our page-anchor and human-in-the-loop mechanisms are the technical realization of this principle in the generative-AI scene---AI assistance is not AI replacement, and the non-delegability of humanistic judgment must be respected by technical architecture. Second, Stapel and Zandhuis (2025), in \textit{Digital Scholarship in the Humanities}, show how Linked Data can model and replicate the knowledge-production process in data-driven humanities research---every decision from primary source to scholarly claim is made explicit \citep{stapel2025linked}. Their work addresses \textbf{data-layer reproducibility}; ours addresses \textbf{generation-layer traceability}; the two paths together constitute the full evidence chain of AI-assisted humanistic research. These are not competing projects but converging efforts toward the same goal: ensuring that AI-assisted humanistic research remains accountable to its sources.

The intersection of existing RAG research, digital humanities, and infrastructure research provides the problem background and dialogue partners for Traceable Scholarship, but a unified normative-infrastructural framework for AI-assisted humanistic research has not yet been formed. This paper seeks to advance the organization of these dispersed requirements into a runnable normative-infrastructural framework, so that source location, knowledge boundaries, refusal to extrapolate, and process review constrain one another along the same system chain. An early version of this framework was publicly released on Zenodo in February 2026 \citep{jing2026zenodo}; this paper integrates the April 2026 Kant case and the June--July 2026 developments in the three-layer architecture and MCP integration on the basis of that conceptual framework.

\section{Conclusion}\label{sec:conclusion}

Generative AI gives humanistic research unprecedented abilities to process, retrieve, organize, and generate materials. But the stronger the capability, the higher the demands on infrastructure and norms. Models can generate fluent explanations, but they cannot automatically provide the conditions under which explanations hold. Humanistic research must require AI outputs to return to materials, pages, editions, processes, and communal review.

This paper proposes Traceable Scholarship as the minimum normative framework for this requirement and presents AIH-Infra as its reference implementation. Contexture preserves pages and structure as documents enter machine processing; Open WebUI AIH-Infra gives knowledge-base Q\&A page-level citation and evidence export; MCP Server lets general agents call these capabilities within explicit knowledge boundaries; and the real intellectual-history case shows that this infrastructure can help researchers expose evidence strength and weakness, correct query strategies, downgrade overstated claims, and form an auditable research process.

Traceability is not a technical detail but the public condition of humanistic research in the age of generative AI. A system that cannot return to source pages, cannot explain knowledge boundaries, cannot expose evidence chains, and cannot admit evidential insufficiency cannot directly bear scholarly judgment, however fluent its explanations. The future of AI-assisted research should not be a smoother black-box authority but a research infrastructure that is more traceable, more reviewable, and more open to counter-evidence and communal questioning.

The framework and implementation in this paper are only a starting point. Next steps include: advancing the acceptance methods of the four-level compliance model from design documents to runnable test suites and completing the three-corpus S2 evaluation described in \S\ref{sec:evaluation}; testing the transferability of traceable workflows in more humanities disciplines (classics, modern history, literary studies, archival studies); standardizing the Evidence/Trace export and intermediate-memo formats so that different research teams can exchange and review one another's process records; and building more formal interoperability interfaces among page anchors, edition metadata, and TEI/IIIF.

Beyond these, we close with three concrete appeals.

\textbf{To tool developers}: Treat page-anchor preservation as a design goal, not a byproduct. Every document-digitization tool, every RAG framework, and every scholarly AI assistant should list Page Anchor Preservation Rate (PAPR) as a core evaluation metric.

\textbf{To humanities scholars}: When adopting AI-assisted research tools, demand page-level traceability. Do not accept ``according to the literature\ldots'' without a source page; treat human verification as part of scholarly responsibility, not an extra burden.

\textbf{To the digital-humanities community}: Establish traceability standards for AI-assisted humanistic research. The four-level compliance model proposed in this paper is a starting point, not an endpoint. We invite scholarly criticism, revision, and extension---especially toward non-textual materials and non-Western scholarly traditions.

Generative AI has built an unprecedented labyrinth for the humanities---fluent, grand, dazzling. But the value of a labyrinth lies not in its spectacle but in whether you can find your way out. Footnotes have endured for five centuries not because they are convenient but because they institutionalize scholarly accountability; page anchors are our generation's attempt to ensure that this accountability continues in the age of generative AI. Ariadne's thread---the thread from AI output back to the source page---must not be cut.

\appendix
\section{Chinese Conceptual Genealogy and Terminological Equivalences}
\label{app:terms}

Several English terms in this paper were stabilized through Chinese-language research notes and infrastructure design work. The table records preferred English terms, acceptable variants, and the limits of translation. The variants are not all interchangeable: the preferred term remains the default in the main text.

\begin{longtable}{@{}p{.19\textwidth}p{.23\textwidth}p{.27\textwidth}p{.25\textwidth}@{}}
\caption{Chinese formulations, English variants, and translation limits}\label{tab:terminology}\\
\toprule
\textbf{Chinese formulation} & \textbf{Preferred English} & \textbf{Acceptable variants} & \textbf{Function and translation limit} \\
\midrule
\cn{可溯源学术研究} & Traceable Scholarship & source-traceable scholarship; traceable scholarly research & The overall normative--infrastructural framework; not equivalent to metadata provenance alone. \\
\addlinespace
\cn{解释外观} & explanatory appearance & appearance that an explanation is already established; apparent explanatory completion & Completed explanation appears before evidential conditions are met; not reducible to hallucination or confabulation. \\
\addlinespace
\cn{语境抹平} & context flattening & contextual flattening; flattening of historical context & Historical differences and material breaks are smoothed away; not ordinary summarization. \\
\addlinespace
\cn{页面锚点／页码锚点} & page anchor & page-level anchor; page locator; page-anchored citation & A stable route from a claim to a page-level source location; a locator is not merely a hyperlink. \\
\addlinespace
\cn{双重页码} & dual page numbers & dual pagination; digital-and-print pagination & Digital-page and print-page coordinates coexist; not duplicate citation. \\
\addlinespace
\cn{引文优先生成／证据先行} & citation-first generation & evidence-first generation; citation-led generation & Evidence constrains the order and strength of fluent explanation; not mechanical footnotes on every sentence. \\
\addlinespace
\cn{证据链／证据回链} & evidence chain & claim--evidence chain; evidence back-link & Links claims, sources, pages, editions, fragments, and process records; citation alone does not guarantee evidential fit. \\
\addlinespace
\cn{当前作用域内无足够证据} & NO\_EVIDENCE & insufficient-evidence state; scoped evidential insufficiency; no-evidence discipline & Explicit insufficiency within the active scope; it does not claim that evidence is absent everywhere. \\
\addlinespace
\cn{人工核验／研究者复核} & human verification & researcher verification; human-led source verification; manual source checking & Researcher-led checking of claim--source fit, pages, editions, and context; not a one-click approval. \\
\addlinespace
\cn{判断降调} & judgment downgrading & claim downgrading; conclusion downgrading; epistemic weakening & The claim is weakened when the evidence warrants a weaker formulation; it is not necessarily rejected. \\
\addlinespace
\cn{作用域契约} & Scope Contract & research-scope contract; retrieval-scope contract; knowledge-boundary contract & Records requested/effective scope, filters, empty boundaries, and tool-call limits; not merely access control. \\
\addlinespace
\cn{过程可审计性} & process-auditability & research-process auditability; workflow auditability; procedural traceability & Preserves enough retrieval, reading, exclusion, generation, and judgment steps to reconstruct key claims; it does not expose model internals. \\
\addlinespace
\cn{幻觉洗白} & hallucination laundering & citation laundering; scholarly laundering of AI output & Weak or sourceless output acquires authority through citation packaging; the focus is the propagation process, not only model error. \\
\bottomrule
\end{longtable}

The preferred Chinese name for the overall framework is \cn{可溯源学术研究}. Earlier working materials sometimes used \cn{可追溯学术研究}; that expression is retained only as a historical variant. \cn{页面锚点} and \cn{页码锚点} remain coexisting formulations: the former is preferable when discussing document structure and processing, while the latter is preferable when discussing formal scholarly citation.

\end{document}